\documentclass[conference]{IEEEtran}
\usepackage[dvipsnames]{xcolor}
\usepackage{times}

\usepackage[numbers,sort]{natbib}
\usepackage{multicol}
\usepackage[bookmarks=true, hidelinks]{hyperref}

\usepackage{amsmath,amssymb,amsfonts}

\usepackage{textcomp}
\usepackage{xspace}

\usepackage{graphicx}
\usepackage[inkscapelatex=false]{svg}

\usepackage[noend]{algpseudocode}
\usepackage{algorithm}

\newcommand{\proc}[1]{\textsc{#1}}
\newcommand{\procnew}[1]{\textbf{\textsc{\color{ForestGreen}#1}}}
\newcommand{\kw}[1]{\textbf{#1}}

\newcommand{\True}[0]{\kw{True}}
\newcommand{\False}[0]{\kw{False}}

\newcommand{\lineref}[1]{Line~\ref{#1}\xspace}
\newcommand{\hybref}[1]{\mbox{\autoref{fig:hybrid}~\textbf{#1})\xspace}}

\newcommand{\Api}{${\Pi_p}$\xspace}
\newcommand{\plan}{plan\xspace}
\newcommand{\Plan}{Plan\xspace}

\begin{document}

\title{
Meta-Policy Learning over \Plan Ensembles for \\
Robust Articulated Object Manipulation
}

\author{
\authorblockN{Constantinos Chamzas}
\authorblockA{Rice University\\
Houston, USA \\
\small\texttt{cchamzas@rice.edu}}
\and
\authorblockN{Caelan Garrett}
\authorblockA{NVIDIA Research\\
Seattle, USA \\
\small\texttt{cgarrett@nvidia.com}}
\and
\authorblockN{Balakumar Sundaralingam}
\authorblockA{
NVIDIA Research\\
Seattle, USA \\
\small\texttt{balakumars@nvidia.com}}
\and
\authorblockN{Lydia E. Kavraki}
\authorblockA{
Rice University\\
Houston, USA \\
\small\texttt{kavraki@rice.edu}}
\and 
\authorblockN{Dieter Fox}
\authorblockA{NVIDIA Research\\
Seattle, USA \\
\small\texttt{~dieterf@nvidia.com}}}

\maketitle

\section{Introduction}

Model-based robotic planning techniques, such as inverse kinematics and motion planning, can endow robots with the ability to perform complex manipulation tasks, such as grasping, object manipulation, and precise placement. 
However, these methods often assume perfect world knowledge and leverage approximate world models.
For example, tasks that involve dynamics such as pushing or pouring are difficult to address with model-based techniques \cite{kejia2022} 
as it is difficult to obtain accurate characterizations of these object dynamics.
Additionally, uncertainty in perception prevents them populating an accurate world state estimate.

Recent works have shown that integrating learning with model-based planning can address some of these limitations~\cite{haro22a, yang2022sequence, agia2022taps}. Following a similar direction, we integrate a geometric model-based motion planner with learning in a mixture-of-experts fashion to solve manipulation problems where the world model and dynamics are only approximately known.

\begin{figure}
\vspace{-2em}
\center
\includegraphics[width=0.8\linewidth]{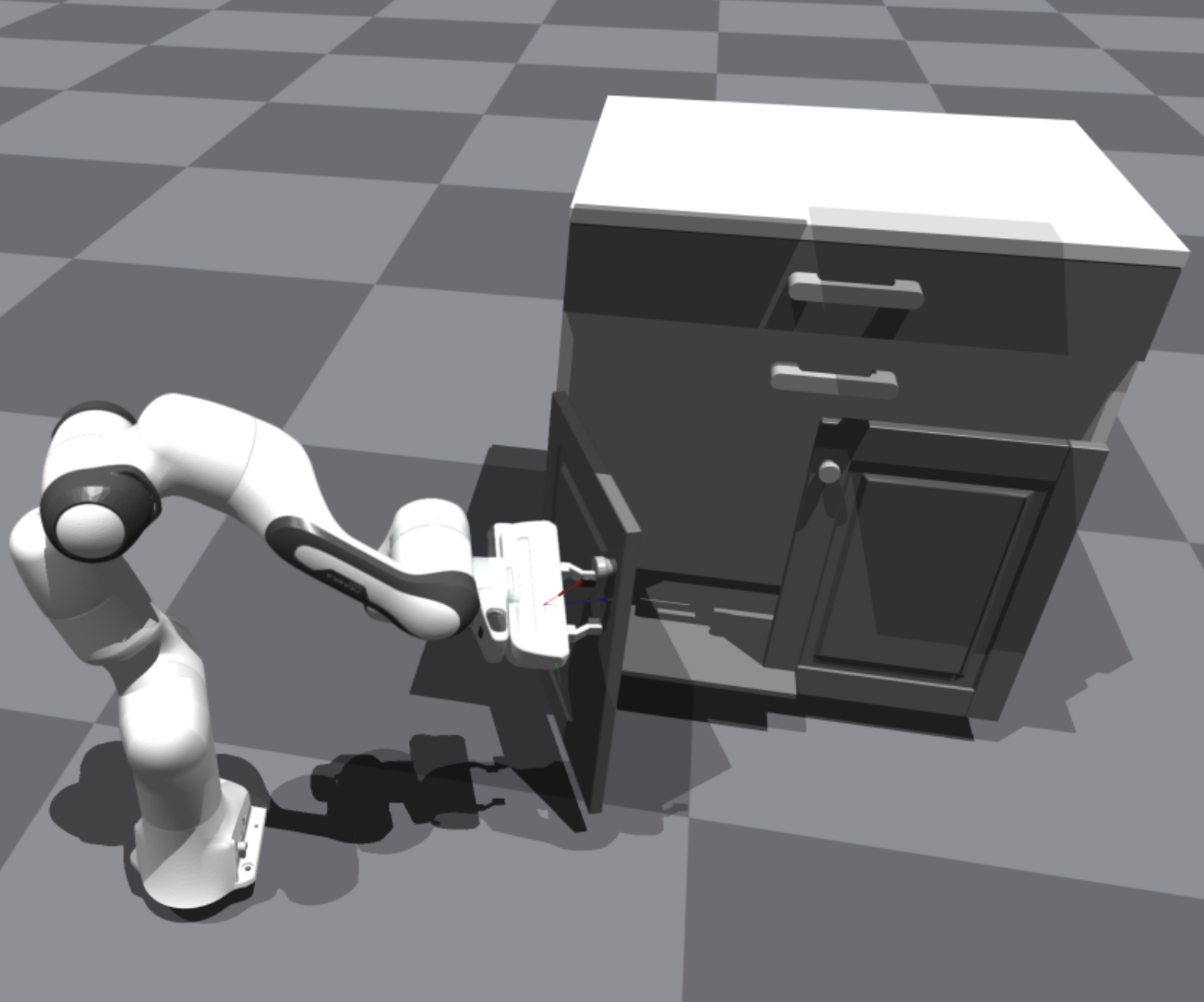}
\caption{Example of the panda robot pushing the left door of the cabinet. In this setting, the robot knows its own state and geometry along with the state and geometry of the cabinet. However, the position of the cabinet is only approximately estimated, and the dynamic properties of the objects are unknown.
In this work, we attempt to solve this task by combining a model-based planner and learning from experience.  
\vspace{-2em}
}
\label{fig:push_door}
\end{figure}

Specifically, we propose using a model-based motion planner to build an {\em ensemble of plans} under different environment hypotheses.
Then, we train a {\em meta-policy} to decide online which plan to track based on the current history of observations.
By leveraging history, this policy is able to switch ensemble plans to circumvent getting ``stuck'' in order to complete the task.
We tested our method on a 7-DOF Franka-Emika robot pushing a cabinet door in simulation, as shown in \autoref{fig:push_door}.  
We demonstrate that a successful meta-policy can be trained to push a door in settings high environment uncertainty, all while requiring little data ($\leq$ 1000 episodes).

{\bf The main contributions of this work are:} 
\begin{enumerate}
    \item An ensemble approach for manipulation policies, which outperforms single-trajectory planners.   
    \item Learning a meta-policy to select among the ensemble.  
    \item Simulated experiments that show our method results in $40\%$ higher success rate than the non-learning baseline.
\end{enumerate}

\section{Related Work}

At the core of model-based planning techniques for robot manipulation lies motion planning \cite{choset2005principles}.
Such techniques include constrained motion planning~\cite{kingsto2021review}, Multi-Modal Motion Planning \cite{hauser2010multi}~(MMMP), and Task and Motion Planning (TAMP)~\cite{garrett2021}.
The main advantages of these methods are that they are general, {\it i.e.} work with any geometries, and have theoretical guarantees with respect to their models, {\it e.g.} probabilistic completeness~\cite{kavraki1996probabilistic}. However, they depend critically on the fidelity of their models, limiting their effectiveness when there are state estimation errors and unmodeled constraints, {\it e.g.}  dynamic constraints such as limited pushing force or friction between objects and the environment.
Finally, these methods often have long (several seconds) planning times, making it difficult for them to react online in real-time to unexpected events. 

Recent advances in learning have prompted researchers to combine learning with model-based planning to address these limitations.
Some techniques try to learn world models \cite{silver2022inventing, hafner2020mastering, wu2023daydreamer, wang2021learning}, warm start policy-learning with model-based techniques \cite{silver2018residual,johannink2019residual}, or improve planning efficiency \cite{chamzas2022learning, yang2022sequence}.
We also leverage learning to improve the downstream performance of a model-based planner. However, we focus on approximate world models with uncertain object poses and geometries and problems that involve dynamics such as pushing.

\section{Methodology} 
In this work, we aim to endow a robotic agent with advanced  manipulation skills regarding grasping, pushing, and even throwing objects under realistic assumptions.
The realistic assumptions that we consider are that a world model can be engineered but is not entirely accurate.
These inaccuracies could be due to pose uncertainty, lack of dynamic modeling, or approximate geometric representations. 
Our method uses a model-based planner to produce candidate plans using the approximate world model and then uses learning from experience to learn a meta-policy that improves task performance.

We will use the toy problem shown in \hybref{a} to demonstrate the key concepts in this work.
In this toy problem, the robot aims to push the red object to the shaded red goal location. The modeled constraints include collision avoidance and joint limits. Unknown aspects include dynamics, namely how the object moves upon contact with the end-effector. Task failure can occur if the object falls over due to the pushing angle. Uncertainty exists in the object's x-axis position estimation and its exact geometry.

The world state is the prorpioceptive state of the robot along with the potentially uncertain state of the world objects.
For example, in the problem shown in \hybref{a}, the state includes the positions of the robot's five joints and an approximate estimate of the position and geometry of the red object along the x-axis.
We consider such state estimation realistic as robots usually have accurate joint encoder measurements but perceive the objects in the world around them with some uncertainty.   

The proposed approach is composed of two main components.
First, a geometric model-based planner creates an ensemble of plans as described in \autoref{subsec:pathpolicy} and shown in \hybref{b}.
Second, given the currently observed state, we learn a meta-policy that chooses plans from the ensemble as described in \autoref{subsec:metapolicy2} and shown in \hybref{b} (the large black arrows).
The proposed hybrid approach leverages the approximate world model to create \plan-ensembles that satisfy known constraints, such as geometric-constraints. Then we use learning to account for unmodeled constraints, {\it e.g.} dynamics, and be robust to estimation errors.

\begin{figure}
\includegraphics[width=\linewidth]{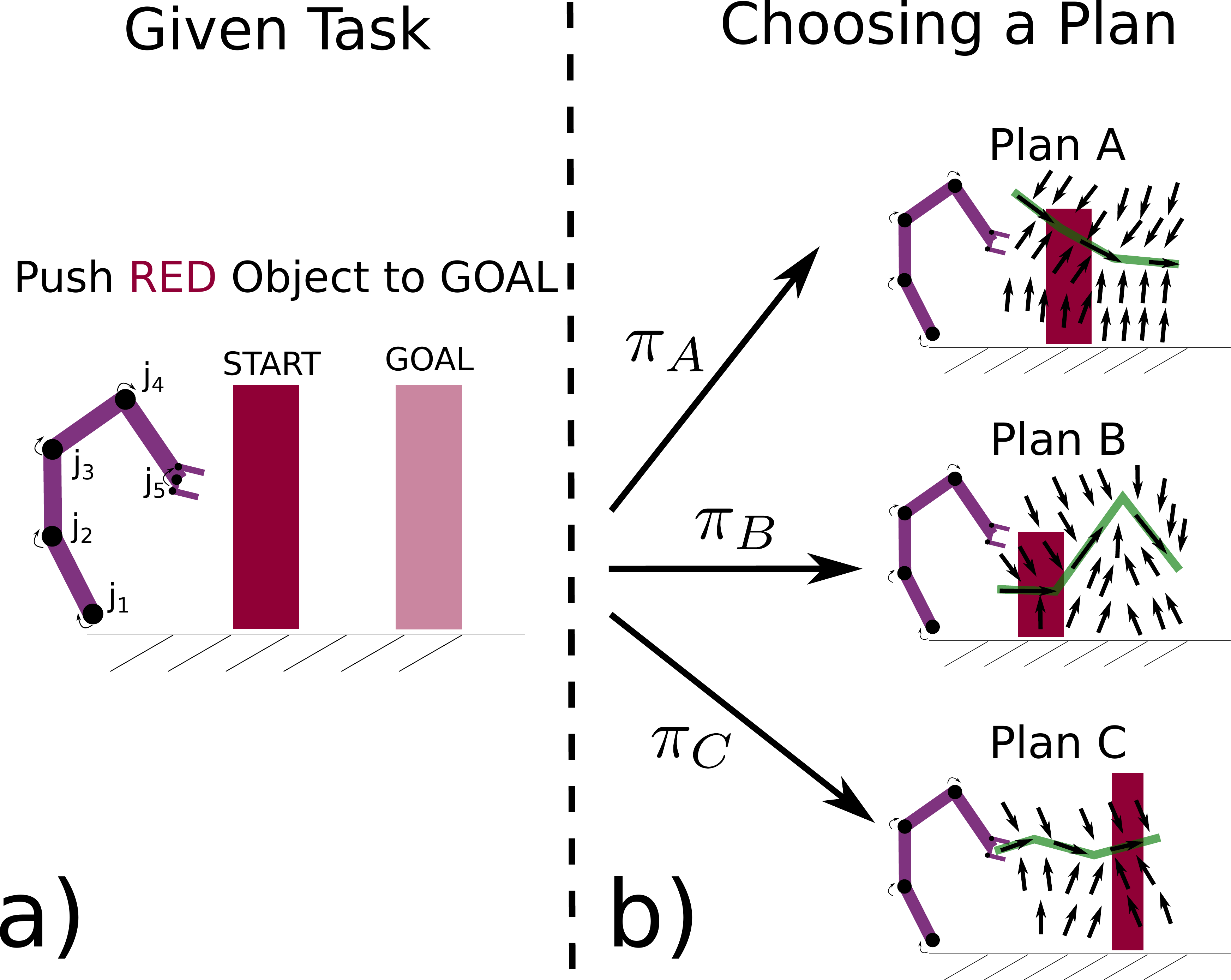}
\vspace{-1.5em}
\caption{
\textbf{a)} 
An illustrative example of a task that combines geometric, dynamic reasoning and uncertainty.    
The robot is tasked with pushing the red object to the goal location, without toppling it over. The robot can control its 5 joints and  estimate within some error the position and geometric height of the red object.   
\textbf{b)} The proposed method uses model-based planning to create several path policies. The path shown with green is produced with a model-based geometric planner and satisfies the geometric (known) constraints of the problem. 
The black arrows represent the \plan, which is the combination of the path and a position controller. Note that the \plan is shown in the end-effector 2D space but its state space is the 5D space of the joints of the robot.} 
\label{fig:hybrid}
\end{figure}

\begin{algorithm}[!ht]
	\caption{Plan Ensemble Pseudocode \label{alg}}
	\begin{algorithmic}[1] %
        \Procedure{Plan-Ensemble}{$G$, \Api, $K$}  
		\State \Api $\gets \emptyset$   \Comment{Plan-Ensembles} 
		\While{$i \leq N$} \Comment{Create Plan-Ensembles} \label{alg:start1}
		\State $\hat{s} \gets \proc{sample-world}()$ \label{alg:sample} 
        \State $p \gets \proc{plan}(\hat{s}, G)$ \Comment{Plan Geometric Path} \label{alg:plan}
		\State $\pi_p \gets \proc{make-policy}(p)$ \Comment{Create \Plan}
		\label{alg:makepolicy}
		\State \Api $\gets$\Api$\cup~\pi_p$  \label{alg:add} 
		\EndWhile \label{alg:end1}
		\While{$t \leq T$} \label{alg:start2} \Comment{Iteratively Reselect \Plan} \label{alg:start}

        \State $o \gets \proc{observe}()$ \label{alg:observe1}
	
        \State $\pi_p^{chosen}\gets \underset{\pi_p \in \Pi_p}{\mathrm{argmax}}~\procnew{best}(o, \pi_p) $  \label{alg:best} 
		\While{$\procnew{progress}(o,p)$} \label{alg:stuck}
		\State $o \gets \proc{observe}()$ \Comment{Observe State} \label{alg:observe2}
      	\If{$o \in G$} 
    		\State \Return \True \Comment{Success!}
    	\EndIf
		\State $r \gets  \proc{execute}(\pi^{chosen}_p)$
            \EndWhile
		\EndWhile \label{alg:end2}
		\State \Return \False \Comment{Failure}
		\EndProcedure
	\end{algorithmic}
\end{algorithm}

\subsection{Create \Plan-Ensembles}\label{subsec:pathpolicy}

Creating \plan-ensembles is described in \lineref{alg:start1} to \lineref{alg:end1} of \autoref{alg}.
In \lineref{alg:sample}, \proc{sample-world()} samples a possible world state according to a given uncertainty distribution.
In \lineref{alg:plan}, given the currently estimated state of the world $\hat{s}$ and a goal $G$, a model-based planner computes a candidate path $p$.  
A path $p$ is as sequence of waypoints where each waypoint specifies the position of the robot and other known movable objects in the world.
The first waypoint is the current state, while the last waypoint must satisfy the goal specification. 
Examples of paths are the green lines shown in \hybref{b}. 

In \lineref{alg:makepolicy}, the path $p$ is converted into a \plan $\pi_p$. 
The path is time-parameterized and a waypoint-tracking trajectory controller is used to follow it. In this context, 
a plan is converted into an atomic
policy that provides an action for any state.     
Example plan actions are shown as the black arrows that attract towards the green path $p$ in \hybref{b}.

Note that each \plan is computed for an estimation of the world state $\hat{s}$.
This is illustrated in \hybref{b}, where the objects for each computed path policy
are either in different locations (inaccurate pose estimation) or have different geometry (inaccurate geometry estimation). Finally, in \lineref{alg:add}, the \plan is added in the plan-ensemble \Api.     

\subsection{Selecting Plans} 
\label{subsec:metapolicy2}

\begin{figure*}[ht!]
\includegraphics[width=\linewidth]{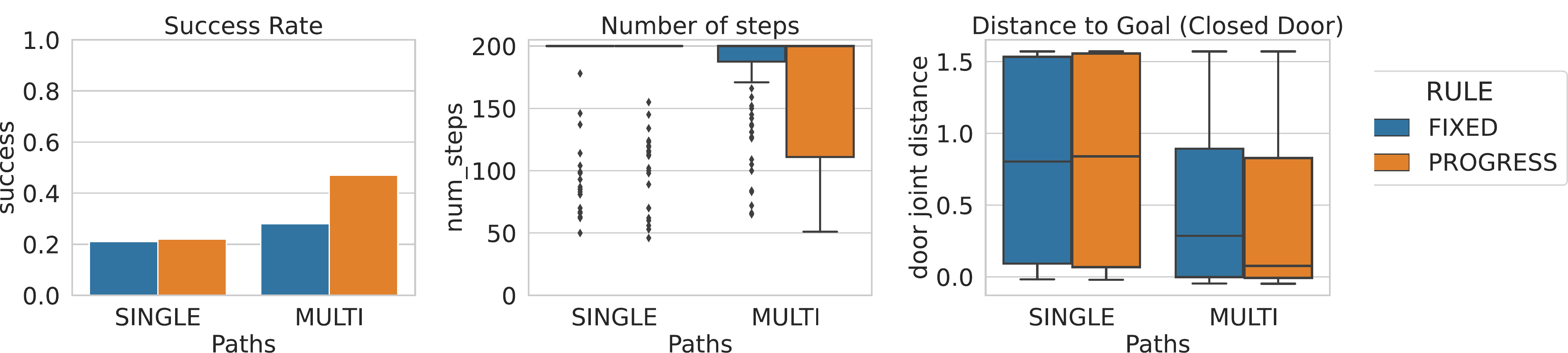}
\vspace{-1em}
\caption{\textit{Success Rate}, \textit{Number of Steps}, and \textit{Distance to Goal} different configurations of method.
All configurations were tested on a 100 different environments, with a maximum of 200 timesteps. \textsc{single} refers to using just a single path policy while \textsc{multi} means that 10 path-policies were used for the path-policy set \Api respectively.
\textsc{fixed/progress} denotes whether a fixed interval or the \procnew{progress} rule was used to choose a new path-policy.}
\vspace{-1.5em}
\label{fig:ablation}
\end{figure*}

Selecting and executing a plan is described in \lineref{alg:start2} to \lineref{alg:end2} of \autoref{alg}. 
Given a plan-ensemble \Api, we need to choose the most appropriate plan to follow for the current observed state $o$.
In this context, the observed state includes the same information as each waypoint of the path.  
We choose a simple supervised approach to learn how to select a plan.

In \lineref{alg:observe1}, the known state of the robot and the world is observed, which can be used to determine if we have reached a terminal state.
An episode terminates when the goal is reached or when maximum number of steps has passed.
An important implementation choice is the decision-frequency at which a new \plan is chosen. 
One end of the spectrum would be choosing a single plan and following it until the end of the episode.
The opposing end would be choosing a new plan at every timestep. 
In this work, we propose using a simple rule that leverages the computed path $p$ to decide when to switch.
Taking advantage of this knowledge, in \lineref{alg:stuck}, the \procnew{progress} function monitors if progress towards the planned path $p$ is made and returns \textit{False} if no progress has been made in the last $k$ timesteps. 
When this function returns \textit{False}, a new plan is chosen, as shown in \lineref{alg:best}.

To generate training data, we first create the \plan-ensemble \Api and then execute plans $\pi_p$ randomly.
After execution, we collect the state, action, and progress triplet and then store it to later train the network \procnew{best}.
We define the distance to goal that was covered by the execution of the plan as progress. 
We use the collected data to train the network in supervised manner.
The network learns to predict the progress that will be achieved, if we execute the plan $\pi^{chosen}_p$ from the current observed state $o$.
During testing, we use the trained network \procnew{best} \lineref{alg:best} to choose the most promising \plan and execute it.
Execution continues until the goal is reached or the \procnew{progress} returns \textit{False}.

\section{Experiments}

We applied our framework to a pushing a cabinet door problem, as shown in \autoref{fig:push_door}. 
The robot task is to push the left cabinet door from the open state to the closed state.    

\subsection{Experiment Setup}
The initial state uncertainty we consider lies in the position of the cabinet relative to the robot. Thus, the \proc{sample-world()} function samples a random position of the cabinet relative to the robot according to the given uncertainty distribution.
Here, we only consider translational uncertainty.  

Given the random sample from this distribution $\hat{s}$, the planner produces a geometric path $p$ that maintains contact perpendicular to the door and intends to push it until it closes.
The pushing contact location on the cabinet door is chosen randomly; different contact points are sampled for each computed path. 
We use TRAC-IK~\cite{tracik} combined with a \textsc{pid} controller to convert the geometric path $p$ to a policy which we refer to as a   \plan $\pi_p$.
To represent the geometric world and generate the \plan-ensembles \Api, we use PyBullet~\cite{coumans2015bullet}. 
We use IsaacGym \cite{makoviychuk2021isaac} for physics and control simulation.

We conducted 2 experiments to examine the performance of our proposed method.
The first experiment(\autoref{subsec:ablation}), studies the  choices of the switching frequency rule and the use of ensembles. 
Specifically, we investigate the effect of the \procnew{progress} rule, and choosing to use a plan-ensemble instead of a single plan . 
The second experiment (\autoref{subsec:metapolicy}), benchmarks the proposed method with respect to a non-learning baseline and ablates different input features to the \procnew{best} neural network.
We evaluated with these metrics:
\begin{itemize}
    \item \textbf{Distance to goal (Progress)}:
    This simply measures how much the door closed.
    A distance of 0 means that the door closed ({\it i.e.} the goal was reached), and a distance of $\pi/2$ means that the door did not move at all. 
    \item \textbf{Success rate}: The percentage of tasks solved successfully within the given horizon.    
    The task is considered successful if the door is closed up to some error threshold.     
    \item \textbf{Number of steps}: The number of timesteps to complete the episode.
    The episode terminates if the door is closed or if the maximum of 200 timesteps is reached.
\end{itemize}

\begin{figure*}[!ht]
\includegraphics[width=\linewidth]{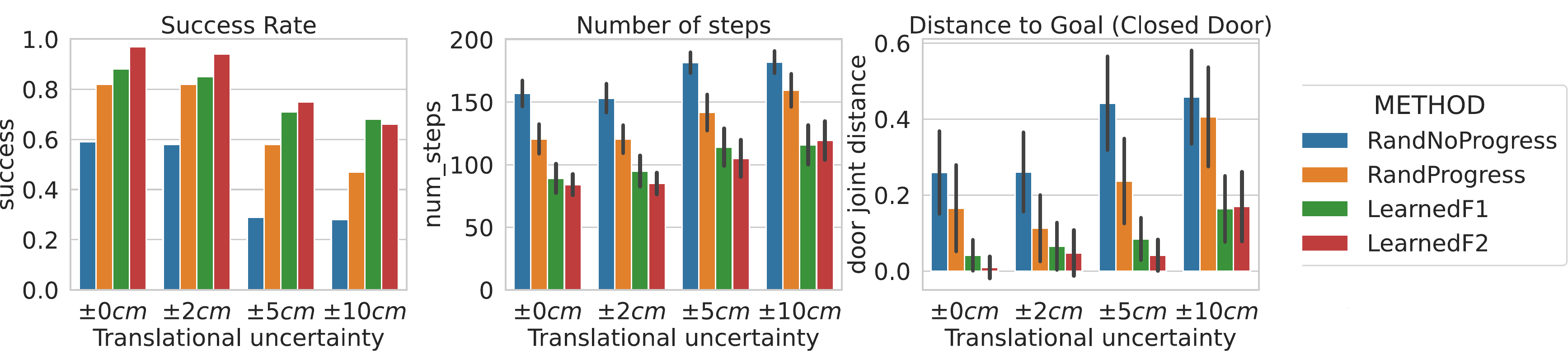}
\vspace{-2em}
\caption{The 1) Success Rate, 2) Number of Steps, and 3) Distance to Goal for both the two random baselines, with and without the \procnew{progress} (RandNoProgress/RandProgress) rule, as well as the two different feature sets (LearnedF1/LearnedF2) used for training the \procnew{best} network. The x-axis denotes the translational estimation error on the cabinet position. The error bars indicate 0.25 of the standard deviation.}
\vspace{-2em}
\label{fig:ranking}
\end{figure*}

\subsection{Using Plan Ensembles and Progress Rule} %
\label{subsec:ablation}

Here, the uncertainty was emulated as a uniform distribution $\pm$10 $cm$ on the position of the cabinet and applied across x,y,z dimensions independently. 
We tested the following:

\begin{enumerate}
    \item \textsc{single/multi} paths: \textsc{single} means that only one plan was used, while \textsc{multi} means that we had a plan-ensemble of 10 plans.
    The plan to follow at each time-step was chosen randomly.
    This comparison motivates why using multiple plans(an ensemble) is better than a single one. 

    \item \textsc{fixed/progress} rule: \textsc{fixed} means that the decision to switch plan was every 10 fixed timesteps, and \textsc{progress} means that the rule described in \autoref{subsec:metapolicy} was used to determine the switching frequency.
    In both cases, the new plan $\pi_p$ is chosen randomly. 
   For \textsc{single}, \textsc{fixed} has no effect, while \textsc{progress} added noise $\pm$ 10cm in the end-effector space to help unstuck the robot.   
\end{enumerate}

The results demonstrate that the \procnew{progress} rule and using an ensemble significantly improves  performance.
Intuitively as long as progress is made along the path, the current plan is good and it should not be switched. The advantage of using an ensemble of plans lies in the fact that part of one plan could apply for part of the episode and part of another plan can apply to another part of the episode.
Thus an ensemble can complete tasks that a single plan could never complete.
The \textsc{multi/progress} combination achieves both the best  distance to goal and also requires the least number of steps to complete the task.
The main takeaways from these results are first that the \procnew{progress} rule helps significantly in all uncertainty settings. 

\subsection{Learning a Meta Policy} %
\label{subsec:metapolicy}
In this experiment, we investigated the improvements in task performance of learning the subroutine \procnew{best}.
First, we use a random strategy to choose and execute path-policies for the pushing task in \autoref{fig:push_door} using the same setting as the experiment setup of \autoref{subsec:ablation} with \textsc{multi} paths and the \procnew{progress} rule enabled.
We collected data from 1000 episodes.

We trained a regressor neural network \procnew{best}, which takes as input the plan $\pi_p$ as well as the current observation $o$ and possibly the history of observations and predicts the progress toward the goal after executing $\pi_p$. 
The neural network architecture has three fully connected layers with batch normalization\cite{ioffe2015batch}.
Each hidden layer has 64 units, and the network was trained to minimize Mean Squared Error (MSE).

An important decision is how to represent the path and history of observations as the input to the neural network. 
We use the following values are input features:

\begin{itemize}
    \item \textit{Current End-Effector(EE) State}: the position, expressed in Cartesian coordinates (3 dim) and the orientation expressed as a quaternion~(4 dim). 
    \item \textit{Current Robot State}: joint values of the robot~(7 dim).
    \item  \textit{Current Door State}: the joint value of the door~(1 dim).
    \item \textit {Final End-Effector State}~(7 dim).    
    \item \textit {Next Immediate End-Effector State}~(7 dim).
    \item  \textit {Midpoint End-effector State}: the midpoint  between Current and Final End-Effector States~(7 dim), using spherical linear orientation interpolation. 
    
 \end{itemize}

The observed state of the world includes the current end effector state, robot joint state, and door state~(15 dim). The first set of features $F1$ includes the observation, the next end-effector goal, and the end effector goal~(29 dim). 
The second set of features $F2$ extends $F1$ by adding the midpoint end-effector state, the difference between the current end-effector state and the next end-effector state, and the difference between the current end-effector state and the final end-effector state. Additionally, the last 5 observations are included~(175 dim).

We trained the neural network with the same data from $\pm 10$ cm uncertainty and tested it in 100 environments under $\pm 0, \pm 2, \pm 5, \pm 10$ cm of translational uncertainty.
We also tested a random strategy similar to the previous section when using and not using the progress rule.
In \autoref{fig:ranking}, the success rate, number of steps included, and distance to goal are shown for the four uncertainty settings.

The learned strategies are better than the random strategy, and the F2 set of features seems to help best in the lower uncertainty regime.
We hypothesize that the $\pm10$ cm regime is too uncertain to leverage the observation history or potentially over-fits, while in the lower regime the observation history can help disambiguate the error in position estimation.
Overall, incorporating the observation history helps, and it is possible to learn the meta policy over the \plan-ensemble. 
A video demonstration of the experiments is available \footnote{{\mbox{\url{https://www.dropbox.com/s/8gosmt9e838icqg/RSS2023-LTAMP.mp4?dl=0}}}}.

\section{Conclusion}
In this work, we proposed a method for manipulation that learns a meta policy over plan-ensembles using approximate world models.
In our preliminary results, we have demonstrated the efficacy of the proposed method over simple baselines in a simulated environment where the robot is tasked with pushing the door of cabinet. 
The two main assumptions of the proposed method are that an approximate model of the world is given and a quantification of the uncertainty of the model is also available.
In the future, we would like to apply the proposed method to a real-world setting and investigate the feasibility of these assumptions. 
We would also like to investigate extending to more manipulation skills, multi-modal planning, and image-based predictions.

\bibliographystyle{unsrtnat}
\bibliography{biblio}

\end{document}